\documentclass[letterpaper, 10 pt, journal,twoside]{ieeetran}

\usepackage[utf8]{inputenc}
\usepackage[colorlinks=true,linkcolor=blue,citecolor=green]{hyperref}
\usepackage[dvipsnames]{xcolor}
\usepackage{graphicx}

\usepackage{amsmath, amssymb, amsfonts}
\usepackage{physics}
\usepackage{cancel}
\usepackage{thmtools, thm-restate}

\newtheorem{remark}{Remark}
\newtheorem{lemma}{Lemma}

\newcommand{\x}{\times}
\newcommand{\cB}{\mathcal{B}}
\newcommand{\cC}{\mathcal{C}}
\newcommand{\cF}{\mathcal{F}}
\newcommand{\cL}{\mathcal{L}}
\newcommand{\cN}{\mathcal{N}}
\newcommand{\bA}{\mathbb{A}}
\newcommand{\bE}{\mathbb{E}}

\newcommand{\R}{\mathbb{R}}
\newcommand{\bT}{\mathbb{T}}
\newcommand{\bU}{\mathbb{U}}
\newcommand{\bX}{\mathbb{X}}
\newcommand{\nU}{{n_u}}
\newcommand{\nX}{{n_x}}
\newcommand{\nT}{{n_\theta}}
\newcommand{\nN}{{n_\xi}}
\newcommand{\la}{\left\langle}
\newcommand{\ra}{\right\rangle}
\newcommand{\lb}{\left[}
\newcommand{\rb}{\right]}
\newcommand{\lc}{\left(}
\newcommand{\rc}{\right)}
\def\blue{\color{blue!70!white}}

\makeatletter
\DeclareRobustCommand\widecheck[1]{{\mathpalette\@widecheck{#1}}}
\def\@widecheck#1#2{%
    \setbox\z@\hbox{\m@th$#1#2$}%
    \setbox\tw@\hbox{\m@th$#1%
       \widehat{%
          \vrule\@width\z@\@height\ht\z@
          \vrule\@height\z@\@width\wd\z@}$}%
    \dp\tw@-\ht\z@
    \@tempdima\ht\z@ \advance\@tempdima2\ht\tw@ \divide\@tempdima\thr@@
    \setbox\tw@\hbox{%
       \raise\@tempdima\hbox{\scalebox{1}[-1]{\lower\@tempdima\box
\tw@}}}%
    {\ooalign{\box\tw@ \cr \box\z@}}}
\makeatother

\title{\LARGE \bf
    Learning Stochastic Optimal Policies via Gradient Descent
}

\IEEEoverridecommandlockouts
\renewcommand\IEEEkeywordsname{Keywords}

\author{Stefano Massaroli$^{1,\star}$, Michael Poli$^{2,\star}$, Stefano Peluchetti$^{3}$\\ Jinkyoo Park$^{2}$, Atsushi Yamashita$^{1}$ and Hajime Asama$^{1}$%
\thanks{$^{1}$ \textit{The University of Tokyo},~ {\tt massaroli@robot.t.u-tokyo.ac.jp}}%
\thanks{$^{2}$ \textit{KAIST}, {\tt poli\_m@kaist.ac.kr}}%
\thanks{$^{3}$ \textit{Cogent Labs},~{\tt speluchetti@cogent.co.jp}}%
\thanks{$^{\star}$ equal contribution authors}%
}
\begin{document}
\maketitle

\begin{abstract}
We systematically develop a learning-based treatment of stochastic optimal control (SOC), relying on direct optimization of parametric control policies. We propose a derivation of adjoint sensitivity results for stochastic differential equations through direct application of variational calculus. Then, given an objective function for a predetermined task specifying the desiderata for the controller, we optimize their parameters via iterative gradient descent methods. In doing so, we extend the range of applicability of classical SOC techniques, often requiring strict assumptions on the functional form of system and control. We verify the performance of the proposed approach on a continuous--time, finite horizon portfolio optimization with proportional transaction costs.
\end{abstract}%


\section{Introduction}
In this work we consider the following class of controlled stochastic dynamical systems:
{
\begin{equation}\label{eq:1}
    \dot x_t = f(t, x_t, u_t) + g(t, x_t, u_t)\xi(t)
\end{equation}
with \textit{state} $x_t\in\bX\subset\R^\nX$ and \textit{control policy} $u_t\in\bU\subset\R^\nU$. $\xi(t)\in\R^\nN$ is a stationary {$\delta$-}correlated Gaussian noise, i.e. $\forall t>0~~~\bE[\xi(t)]=0$ and $\forall s, t$ such that  $0<s<t$ it holds $\bE[\xi(s)\xi(t)]={\delta}(s-t)$. The RHS of \eqref{eq:1} comprises a \textit{drift} {term} $f:\R\x\bX\x\bU\rightarrow\R^{\nX}$ and a \textit{diffusion} {term} $g :\R\x\bX\x\bU\rightarrow\R^{\nX\x\nN}$.
}
This paper develops a novel, {systematic approach} to {learning} optimal {control policies} for systems in the form \eqref{eq:1}, with respect to smooth scalar objective function{s}.

{The link between \textit{stochastic optimal control} (SOC) and learning has been explored in the discrete--time case \cite{bertsekas2004stochastic} with policy iteration and value function approximation methods \cite{yu2013q,wu2017policy} seeing widespread utilization in \textit{reinforcement learning} \cite{levine2018reinforcement}. Adaptive stochastic control has obtained explicit solutions through strict assumptions on the class of systems and objectives \cite{tse1972adaptive}, preventing its applicability to the general case. 
Forward--backward SDEs (FBSDEs) have been also proposed to solve classic SOC problems \cite{peng1999fully}, even employing neural approximators for value function dynamics \cite{pereira2019learning, wang2019deep}.
A further connection between SOC and machine learning has also been discussed within the \textit{continuous--depth} paradigm of neural networks \cite{weinan2017proposal,chen2018neural,massaroli2020dissecting,massaroli2020stable}. \cite{peluchetti2020infinitely} e.g. showed that fully connected residual networks converge, in the infinite depth and width limit, to diffusion processes.}

A different class of techniques for SOC involves the analytic \cite{arnold1974stochastic} or approximate solution \cite{kushner2001numerical} of Hamilton--Jacobi--Bellman (HJB) {optimality conditions}. {These approaches either restrict the class of objectives functions to preserve analytic tractability, or develop specific approximate methods which generally become intractable in high--dimensions.}

Here, we explore a different direction, motivated by the affinity between neural network training and optimal control which involve the reliance on carefully crafted objective functions encoding task--dependent desiderata.

In essence, the \textit{raison d'\^etre} of synthesizing optimal stochastic controllers through gradient--based techniques is to enrich the class of objectives for which an optimal controller can be found, as well scale the tractability to high-dimensional regimes. This is in line with the empirical results of modern machine learning research where large deep neural networks are often optimized on high--dimensional non--convex problems with outstanding performance \cite{schmidhuber2015deep,goodfellow2016deep}.
{Gradient--based methods are also being explored in classic control settings to obtain a rigorous characterization of optimal control problems in the \textit{linear--quadratic} regime \cite{furieri2020learning,bu2019lqr}}.
\subsubsection*{Notation}
Let $(\Omega, \cF, P)$ be a probability space. If a property (event) $A\in\cF$ holds with $P(A)=1$, we say that such property holds \textit{almost surely}. A family of $\bX$--valued random variables {defined on a compact time domain $\bT\subset \R$} $\{x_t\}_{t \in\bT}$ is called \textit{stochastic process} and is \textit{measurable} if $x_t(A)$ is measurable with respect to the $\sigma$-algebra $\cB(\bT)\x\cF$ being $\cB(\bT)$ the Borel-algebra of $\bT$. As a convention, we use $\int_s^t = -\int_t^s$ if $s<t$ and we denote with $\delta$ the Krocnecker delta function.

\section{Stochastic Differential Equations}
Although \eqref{eq:1} ``\textit{looks like a differential equation, it is really a meaningless string of symbols''} \cite{van1981ito}. This relation should be hence treated {only as} a \textit{pre--equation} and cannot be studied in this form. {Such} ill--posedness arises from the fact that, being $\xi{(t)}$ a $\delta$--autocorrelated process, the noise fluctuates an infinite number of times with infinite variance\footnote{From a control theoretical perspective, if $\xi(t)\in\R$ is, for instance, a white noise signal, its energy $\int_{-\infty}^\infty |F(\xi)(\omega)|\dd\omega$ would not be finite ($F(\cdot)$ denotes the Fourier transform)} {in any time interval}. Therefore, a rigorous treatment of the model {requires a different \textit{calculus} to consistently interpret the integration of the RHS of \eqref{eq:1}}. The resulting well-defined version of \eqref{eq:1} is known as a \textit{Stochastic Differential Equation} (SDE) \cite{oksendal2003stochastic}.
\subsection{It\^o--Stratonovich Dilemma} According to Van Kampen \cite{van1981ito}, $\xi(t)$ might be thought as a random sequence of $\delta$ functions causing, at each time $t$, a sudden jump of $x_t$. The controversies on the interpretation of \eqref{eq:1} arises over the fact that it does not specify which value of $x$ should be used to compute $g$ when the $\delta$ functions {are applied}. There exist two main interpretations of the issue, namely It\^o's \cite{ito1942differential} and Stratonovich's \cite{stratonovich1966new}. It\^o prescribes that $g$ should be computed {with the value of $x$} \textit{before} the jump while Stratonovich uses the {\textit{mean value}} of $x$ before and after the jump. This {choice} leads to two different (yet, both admissible {and equivalent}) types of integration. Formally, {consider a compact time horizon $\bT=[0, T],~T>0$ and} let $\{B_t\}_{t\in\bT}$ be the standard $\nN$-dimensional Wiener process defined on a filtered probability space $(\Omega, \cF, P;\{\cF_t\}_{t\in\bT})$ which is such that $B_0=0$, $B_t$ is almost surely continuous in $t$, nowhere differentiable, has independent Gaussian increments, namely $\forall s, t\in\bT~s<t\Rightarrow B_t-B_s\sim \cN(0,t-s)$ and, {for all $t\in\bT$, we have} $\xi(t)\dd t = \dd B_t$. Moreover, let $\phi_t:=g(t, x_t, u_t)$. It\^o and Stratonovich integral calculi are then defined as $\int_0^T \phi_t\dd B_t =\lim_{|D|\rightarrow 0}\sum_{k=1}^K\phi_{t_{k-1}}(B_{t_k}-B_{t_{k-1}})$ and $\int_0^T \phi_t{\blue\circ}\dd B_t=\lim_{|D|\rightarrow 0}\frac{1}{2}\sum_{k=1}^K \left(\phi_{t_k}-\phi_{t_{k-1}}\right)(B_{t_k}-B_{t_{k-1}})$, respectively,
where $D$ is a given partition of $\bT$, $D:=\{t_k:0=t_0<t_1<\cdots<t_K=T\}$, $|D|=\max_k(t_k-t_{k-1})$ and the limit is intended in the mean--square sense, (see \cite{kunita2019stochastic}) if exists. Note that the symbol ``$\circ$'' in $\circ \dd B_t$ is used only to indicate that the integral is interpreted in the Stratonovich sense and does not stand for function composition. In the Stratonovich convention, we may use the standard rules of calculus\footnote{see e.g. It\^o's Lemma in Stratonovich form \cite[Theorem 2.4.1]{kunita2019stochastic}, where the \textit{chain rule} of classic Riemann--Stieltjes calculus is shown to be preserved under the sign of Stratonovich integrals.} while this is not the case for It\^o's. This is because Stratonovich calculus corresponds to the limit case of a smooth process with a small finite auto--correlation approaching $B_t$ \cite{wong1965convergence}.  
Therefore, there are two different interpretations of \eqref{eq:1}:
$\dd x_t = f(t, x_t, u_t)\dd t + g(t, x_t, u_t)\dd B_t$ and $\dd x_t = f(t, x_t, u_t)\dd t + g(t, x_t, u_t)\circ\dd B_t$.
Despite their name, SDEs are formally defined as integral equations due to the non-differentiability of the {Brownian} paths $B_t$.

From a control systems perspective, on one hand Stratonovich interpretation seems more appealing for noisy physical systems as it captures the limiting case of multiplicative noise processes with a very small finite auto--correlation, often a reasonable assumption in real applications. However, if we need to treat diffusion terms as intrisic components of the system's dynamics or impose martingale assumptions on the solutions of the SDE, the It\^o convention should be adopted. In particular, this is always the case of financial applications, where stochastic optimal control is widely applied (see e.g. \cite{merton1969lifetime}).

Within the scope of this paper, we will {mainly} adopt the Stratonovich convention for consistency of notation. Nonetheless, all results presented in this manuscript can be equivalently derived for the It\^o case. Indeed an It\^o SDE can be transformed into a Stratonovich one (and viceversa) by the equivalence relation between the two calculi: for all $t\in\bT$ $\int_0^t \phi_s\circ \dd B_s = \int_0^t \phi_s\dd B_s + \frac{1}{2}\int_0^t \sum_{i=1}^\nN \frac{\partial \phi_s^{(i)}}{\partial x}\phi_s^{(i)}\dd s$.
\subsection{Stratonovich SDE} We will formally introduce Stratonovich SDEs following the treatment of Kunita \cite{kunita1997stochastic, kunita2019stochastic}. For $0\leq s\leq t\leq T$ we denote by $\cF_{s,t}$ the smallest $\sigma$-algebra containing all null sets of $\cF$ with respect to which, {for all $v,w\in\bT$ such that $s\leq v\leq w\leq t$, $B_w-B_v$ is measurable}. $\{\cF_{s, t}\}_{s\leq t;s,t\in\bT}$ is called a \textit{two--sided filtration} generated by the Wiener process and we set write $\cF_t = \cF_{0,t}$ for compactness. Then, $\{\cF_t\}_{t\in\bT}$ is itself a filtration with respect to which $B_t$ is $\cF_t$--adapted. Further, we let $f, g$ to be {bounded in $\bX$, infinitely differentiable in $x$, continuously differentiable in $t$, uniformly continuous in $u$ and we assumme} the controller $\{u_t\}_{t\in\bT}$ to be a $\cF_t$--adapted process. Given an initial condition $x_0\in\bX$ assumed to be a $\cF_0$ measurable random variable, we suppose that there exists a $\bX$-valued continuous $\cF_t$-adapted semi--martingale $\{x_t\}_{t\in\bT}$ such that 
\begin{equation}\label{eq:2}
    x_T = x_0 + \int_0^T f(t, x_t, u_t)\dd t + \int_0^T g(t, x_t, u_t)\circ\dd B_t,
\end{equation}
almost surely. Path--wise existence and uniqueness of solutions, i.e. if, for all $t\in\bT$, two solutions $x_t,~x_t'$ such that that $x_0=x_0'$ satisfy $\forall t\in\bT ~x_t=x_t'$ almost surely, is guaranteed under {our class assumptions on $f, g$ and the process $\{u_t\}_{t\in\bT}$}\footnote{Note that less strict sufficient conditions only required uniform Lipsichitz continuity of $f(t, x_t, u_t) + \frac{1}{2}\sum_{i=0}^{\nN}g^{(i)}(t, x_t, u_t) [{\partial g^{(i)}(t, x_t, u_t)}/{\partial x}]$ and $g(t, x_t, u_t)$ w.r.t. $x$ and uniform continuity w.r.t. $t$ and $u$.}.

If, as assumed here, $f$, $g$ are functions of class $\cC^{1,\infty}$ in $(x_t,t)$ uniformly {continuous} in $u$ {with bounded derivatives w.r.t $x$ and $t$}, and $\{u_t\}_{t\in\bT}$ belongs to some admissible control set $\bA$ of $\bT\rightarrow\bU$ functions, then{, given a realization of the Wiener process}, there exists a $\cC^\infty$ mapping $\Phi$ called \textit{stochastic flow} from $\bX \x \bA$ to the space of {absolutely} continuous functions $[s, t]\rightarrow\bX$ such that 
\begin{equation}\label{eq:3}
    x_t = \Phi_{s}(x_s, \{u_s\}_{s\leq t;s,t\in\bT})(t)~~~s\leq t;s,t\in\bT;x_s\in\bX
\end{equation}
almost surely. For the sake of {compactness we denote the RHS of \eqref{eq:3}} with $\Phi_{s,t}(x_s)$. {It is worth to be noticed that the collection} $\{\Phi_{s, t}\}_{s\leq t;s,t\in\bT}$ satisfies the \textit{flow property} (see \cite[Sec. 3.1]{kunita2019stochastic}) 
$\forall s,t, v\in\bT: s<v<t\Rightarrow\Phi_{v,t}(\Phi_{s,v}(x_s))=\Phi_{s,t}(x_s)$ and that it is also a diffeomorphism \cite[Theorem 3.7.1]{kunita2019stochastic}, i.e. there exists an \textit{inverse flow} $\Psi_{s,t}:=\Phi^{-1}_{s,t}$ which satisfies the \textit{backward} SDE
\[
    \begin{aligned}
        \Psi_{s,t}(x_s) = x_s -&\int_s^t f(v, \Psi_{s,v}(x_s), u_v)\dd v\\[-3pt]
        &-\int_s^t g(v, \Psi_{s,v}(x_s), u_v)\circ\dd \widecheck B_v.
    \end{aligned}
\]
{being $\{\widecheck B_t\}_{t\in\bT}$ the realization of the \textit{backward} Wiener process defined as  $\widecheck B_t := B_t - B_T$ for all $t\in\bT$}. The diffeomorphism property of $\Phi_{s,t}$ {thus yields:} $\Psi_{s,t}(\Phi_{s,t}(x_s))=x_s$.

{Therefore, it is possible to \textit{reverse} the solutions of Stratonovich SDEs in a similar fashion to \textit{ordinary} differential equations (ODEs)} by storing/generating identical realizations of the noise process. From a practical point of view, under mild assumptions on the chosen numerical SDE scheme, approximated solutions $\hat{\Phi}_{s,t}(x_s)$ satisfy $\forall x_s\in\bX ~~\hat{\Psi}_{s,t}(\hat{\Phi}_{s,t}(x_s))\rightarrow x_s$ in probability as the discretization time-step $\epsilon\rightarrow 0$ \cite[Theorem 3.3]{li2020scalable}.
%

\section{Direct Stochastic Optimal Control}
We consider the optimal control problem for systems of class $\eqref{eq:1}$ interpreted in the Stratonovich sense. In particular, we aim at determining a control process $\{u_t\}_{t\in\bT}$ within an admissible set of functions $\bA$ to minimize or maximize in expectation some scalar criterion $J_u$ of the form
\begin{equation}\label{eq:4}
        J_u(x_0) = \int_\bT \gamma(t, \Phi_{0,t}(x_0), u_t)\dd t  + \Gamma(\Phi_{0,T}(x_0), u_T),
\end{equation}
measuring the performance of the system. Within the scope of the paper, we characterize $\bA$ as set of all control processes $\{u_r\}_{t\in\bT}$ which are $\cF_t$--adapted, have values in $\bU$ and such that $\bE[\int_\bT\|u_t\|_2^2]\dd t\le\infty$. {To emphasize the (implicit) dependence of $J_u(x_0)$ on the realization of the Wiener process, we often explicitly write $J_u(x_0, \{B_t\}_{t\in\bT})$}.
\subsection{Stochastic Gradient Descent}
In this work we propose to directly seek the optimal controller via mini-batch \textit{stochastic gradient descent} (SGD) (or \textit{ascent}) optimization \cite{robbins1951stochastic, ruder2016overview}. The algorithm works as follows: given a scalar criterion $J_\alpha(x_0,  \{B_t\}_{t\in\bT})$ dependent on the variable $\alpha$ and the Brownian path, it attempts at computing $\alpha^*=\arg \min_{\alpha} \mathbb{E}[J_\alpha(x_0, \{B_t\}_{t\in\bT})]$ or $\alpha^*=\arg \max_{\alpha} \mathbb{E}[J_\alpha(x_0, \{B_t\}_{t\in\bT})]$ starting from an initial guess $\alpha_0$ by updating a candidate solution $\alpha_k$ recursively as $\alpha_{k+1} = \alpha_{k} \pm \frac{\eta_k}{N}\sum_{i=1}^N\frac{\dd}{\dd\alpha} J_{\alpha_k}(x_0, \{B_t\}_{t\in\bT}^i),$
where $N$ is a predetermined number of independent and identically distributed samples $\{B_t\}_{t\in\bT}^{i}$ {of the Wiener process}, $\eta_k$ is a positive scalar \textit{learning rate} and the sign of the update depends on the minimization or maximization nature of the problem. If $\eta_k$ is suitably chosen and $J_\alpha$ is convex, $\alpha_k$ converges in expectation to the minimizer of $J_\alpha$ as $k\rightarrow\infty$. Although global convergence is no longer guaranteed in the non--convex case, SGD--based techniques have been employed across application areas due to their scaling and unique convergence characteristics. These methods have further been refined over time to show remarkable results even in non--convex settings \cite{de2015global,allen2016variance}. 
\subsection{Finite Dim. Optimization via Neural Approximators}
Consider the case in which the criterion $J_u$ has to be minimized. Locally optimal control processes $u = \{u_t\}_{t\in\bT}\in\bA$ could be obtained, in principle, by iterating the SGD algorithm for the criterion $J_u$ in the set $\bA$ of admissible control processes. Since $\bA$ is at least in $L_2(\bT\rightarrow\bU)$ we could preserve local convergence of SGD in the function-space \cite{smyrlis2004local}. An idealized function--space SGD algorithm would compute iterates of the form
\vspace*{-1mm}
\begin{equation*}
        u^{k+1} = u^{k} - \frac{\eta_k}{N}\sum\limits_{i=1}^N\dfrac{\delta}{\delta u} J_u(x_s0, \{B_t\}_{t\in\bT}^i)
\end{equation*}
where $u^k$ is the solution at the $k$th SGD iteration, $\delta/\delta (\cdot)$ is the variational or functional derivative and $J_u$ satisfies $\delta\bE[J_u(x_s,\{B_t\}_{t\in\bT})]/\delta u= \bE[\delta J_u(x_s,\{B_t\}_{t\in\bT})/\delta u]$.

At each \textit{training step} of the controller, this approach would thus perform $N$ independent path simulations, compute the criterion $J_u$ and apply the gradient descent update.
While local convergence to optimal solutions is still ensured under mild regularity and smoothness assumptions, obtaining derivatives in function space turns out to be computationally intractable. Any infinite-dimensional algorithm needs to be discretized in order to be practically implementable. We therefore seek to reduce the problem to finite dimension by approximating $u^k_t$ with a \textit{neural network}\footnote{Here by \textit{neural network} we mean any \textit{learnable} parametric function $u_\theta$ with some approximation power for specific classes of functions.}.

Let $u_{\theta, t}:\theta,t\mapsto u_{\theta, t}$ be a neural network with parameters $\theta\in\R^{\nT}$ which we use as {functional approximators for} candidate optimal controllers, i.e. $u_{\theta, t} \approx u_t^*~\forall t\in\bT$. Further, {we denote} $f_\theta(t, x_t) := f(t, x_t, u_{\theta, t})$ and $g_\theta(t, x_t) := g(t, x_t, u_{\theta, t})$ {and the optimization criterion}
\begin{equation}\label{eq:5}
    J_\theta(x_0) = \int_\bT \gamma_\theta(t, \Phi_{s,t}(x_0))\dd t  + \Gamma_\theta(T,\Phi_{s,T}(x_0))
\end{equation}
with $\gamma_\theta(\cdot, \cdot) = \gamma(\cdot, \cdot, u_{\theta,t})$ and $\Gamma_\theta(\cdot, \cdot) = \Gamma(\cdot, \cdot, u_{\theta,T})$. Then, the optimization problem turns into finding via gradient descent the optimal parameters $\theta$, by iterating
\begin{equation}\label{eq:SGD}
        \theta^{k+1} = \theta^{k} - \frac{\eta_k}{N}\sum\limits_{i=1}^N\dfrac{\dd}{\dd\theta}J_{\theta^k}(x_0, \{B_t\}_{t\in\bT}^i).
\end{equation}

If strong {costraints} over the set of admissible controllers $\bA$ are imposed, the approximation problem can be rewritten onto a complete orthogonal basis of $\bA$ and { $u_\theta$ is parameterized by a truncation of the eigenfunction expansion, rather than a neural network}.

\begin{remark}[Heuristics]
    As common practice in machine learning, the proposed approach relies heavily on the following empirical assumptions 
    \begin{itemize}
        \item[$i.$] The numerical estimate of the mean gradients will be accurate enough to track the direction of the true $\dd J_\theta/\dd\theta$. Here, estimation errors of the gradient will be introduced by the numerical discretization of the SDE and the finiteness of independent path simulations.
        \item[$ii.$] The local optima controller reached by gradient descent/ascent will be \textit{good enough} to satisfy the performance requirements of the control problem.
    \end{itemize}
\end{remark}
In order to perform the above gradient descent we therefore need to compute the gradient (i.e. the sensitivity) of $J_u$ with respect to the parameters $\theta$ in a computationally efficient manner. In the following we detail different approaches to differentiate through SDE solutions.
\section{Cost Gradients and Adjoint Dynamics}
The most straightforward approach for computing path--wise gradients (i.e. independently for each realization of $\{B_t\}_{t\in\bT}$) is by direction{al} differentiation\footnote{Differentiability under the integral sign follows by our smoothness assumptions} i.e.
\begin{equation*}
    \frac{\dd J_\theta}{\dd \theta} = \int_\bT \left[\frac{\partial\gamma_\theta}{\partial x_t}\frac{\dd x_t}{\dd\theta} + \frac{\partial\gamma_\theta}{\partial \theta}\right]\dd t  + \frac{\partial\Gamma_\theta}{\partial x_T}\frac{\dd x_T}{\dd\theta} + \frac{\partial\Gamma_\theta}{\partial \theta} 
\end{equation*}
where the quantities $\dd x_t/\dd \theta$ and $\dd x_T/\dd \theta$ can be obtained with the following result.
\begin{restatable}[Path--wise Forward Sensitivity]{proposition}{Fw}\label{thm:1}
    Let $S_t = \dd x_t/\dd\theta$. Then, $S_t$ is a $\{\cF_t\}$-adapted process satisfying
    \begin{equation*}
        \dd S_t = \left[\frac{\partial f_\theta}{\partial x_t}S_t + \frac{\partial f_\theta}{\partial \theta}\right]\dd t +  \sum_{i=1}^\nN \left[\frac{\partial g_{\theta}^{(i)}}{\partial x_t}S_t + \frac{\partial g_{\theta}^{(i)}}{\partial\theta}\right]\circ\dd B_t^{(i)}
    \end{equation*}
\end{restatable}
\IEEEproof
    The proof is an extension of the forward sensitivity analysis for ODEs (see  \cite[Sec. 3.4]{khalil2002nonlinear}) to the stochastic case. Given the SDE of interest
    $$
        x_t = x_0 + \int_\bT f_\theta(t, x_t)\dd t+ \int_\bT g_\theta(t, x_t)\circ\dd B_t
    $$
    differentiating under the integral sign w.r.t. $\theta$ gives
    $$
        \begin{aligned}
            \frac{\dd x_t}{\dd\theta} &= \int_\bT\left[\frac{\partial f_\theta}{\partial x_t}\frac{\dd x_t}{\dd\theta} + \frac{\partial f_\theta}{\partial\theta}\right]\dd t\\
            &+  \sum_{i=1}^\nN \int_\bT\left[\frac{\partial g_{\theta}^{(i)}}{\partial x_t}\frac{\dd x_t}{\dd\theta} + \frac{\partial g_{\theta}^{(i)}}{\partial\theta}\right]\circ\dd B_t^{(i)}
        \end{aligned}
    $$
    and result follows setting $S_t = \dd x_t/\dd \theta$.
    That differentiating under the integral sign is allowed follows by our assumptions on $f_\theta$ and $g_\theta$ and by an application of \cite[Lemma 3.7.1]{kunita2019stochastic} to the augmented SDE $(x_t,\theta)$.
\endIEEEproof
The main issue with the forward sensitivity approach is its curse of dimensionality with respect to the number of parameters in $\theta$ and state dimensions $n_x$ as it requires us to solve an $\nX\times\nT$ matrix--valued SDE for the whole time horizon $\bT$. At each integration step the full Jacobians $\partial f_\theta/\partial\theta,~\partial g_\theta/\partial\theta$ are required, causing memory and computational overheads in software implementations. Such issue can be overcome by introducing \textit{adjoint} backward gradients.
\vspace{-1mm}
\begin{restatable}[Path--wise Backward Adjoint Gradients]{theorem}{Adj}\label{thm:2}
    Consider the cost function \eqref{eq:5} and let $\lambda\in\cC^1(\bT\rightarrow\bX)$ be a Lagrange multiplier. Then, $\dd J_\theta/\dd\theta$ is given by
    \begin{equation*}
        \frac{\dd J_\theta}{\dd\theta} = \frac{\partial \Gamma_\theta}{\partial\theta}+\int_\bT\left[\lambda_t^\top\frac{\partial f_\theta}{\partial\theta} + \frac{\partial \gamma_\theta}{\partial \theta}\right]\dd  t + \sum_{i=1}^\nN\int_\bT \lambda_t^\top\frac{\partial g_{\theta}^{(i)}}{\partial\theta}\circ\dd B_t^{(i)}
    \end{equation*}
    where the Lagrange multiplier $\lambda_t$ satisfies the following backward Stratonovich SDE:
    \begin{equation*}
        \begin{aligned}
            \dd \lambda_t^{\top} &= -\left[\lambda_t^\top\frac{\partial f_\theta}{\partial x_t} + \frac{\partial \gamma_\theta}{\partial x_t}\right]\dd  t - \sum_{i=1}^\nN \lambda_t^\top\frac{\partial g^{(i)}_{\theta}}{\partial x_t}\circ\dd {\widecheck{B}_t^{(i)}}\\
            \lambda_T^\top &= \frac{\partial \Gamma_\theta}{\partial x_T}
        \end{aligned}
    \end{equation*}
 \end{restatable}
\vspace{2mm}
\begin{lemma}[Line integration by parts]\label{lem:1}
Let $\bT$ be a compact subset of $\R$ and $r:\bT\rightarrow\R^n$, $v:\bT\rightarrow\R^n$ such that $\forall t\in\bT,~t\mapsto r_t:=r(t)\in\cC^1$ and $t\mapsto v_t:=v(t)\in\cC^1$.
We have $\int_\bT\langle r_t,\dd v_t\rangle = \langle r_T, v_T\rangle - \langle r_0, v_0\rangle - \int_\bT\langle \dd r_t, v_t\rangle$.
\end{lemma}
\IEEEproof
    The proof follows immediately from integration by parts using $\dd r_t=\dot r_t\dd t$, $\dd v_t=\dot v_t\dd t$.
\endIEEEproof
\IEEEproof \textit{(Theorem \ref{thm:2})}
    Let $\cL$ be the Lagrangian of the optimization problem and the process $\lambda_t\in\cC^1(\bT\rightarrow\bX)$ an $\cF_t$-adapted Lagrange multiplier. We consider $\cL$ of the form
    \[
        \begin{aligned}
            \cL &= J_\theta({x_0}) - \int_\bT \la \lambda_t, \dd x_t - f_\theta(t, x_t)\dd t - g_\theta(t, x_t)\circ\dd B_t\ra\\
            &=\Gamma_\theta(x_T) + \int_\bT \gamma_\theta(t, x_t)\dd t \\
            &- \int_\bT\la \lambda_t, \dd x_t - f_\theta(t, x_t)\dd t - g_\theta(t, x_t)\circ\dd B_t\ra
        \end{aligned}
    \]
    Since $\dd x_t - f_\theta(t, x_t)\dd t - g_\theta(t, x_t)\circ\dd B_t=0$ by construction, the integral term is always null, $\dd \cL/\dd \theta = \dd J_\theta(x_0)/\dd \theta$ and the Lagrange multiplier process can be freely assigned. Thus, 
    \begin{equation*}
        \begin{aligned}
            \frac{\dd J_\theta}{\dd \theta}& = \frac{\dd \cL}{\dd\theta} = \frac{\partial \Gamma_\theta}{\partial\theta} + \frac{\partial \Gamma_\theta}{\partial x_T} \frac{\dd x_T}{\dd \theta}+\frac{\dd}{\dd\theta}\int_\bT \gamma_\theta(t, x_t)\dd t\\
            &- \frac{\dd}{\dd\theta} \int_\bT\la \lambda_t, \dd x_t - f_\theta(t, x_t)\dd t - g_\theta(t, x_t)\circ\dd B_t\ra
        \end{aligned}
    \end{equation*}
    Note that, by Lemma \ref{lem:1}, we have $\int_\bT\la\lambda_t, \dd x_t\ra = \la\lambda_t,x_t\ra|_0^T-\int_\bT\la\dd\lambda_t,x_t\ra$. For compactness, we will omit the argument of all functions unless special cases. We have
    \begin{equation}
        \begin{aligned}
            \frac{\dd J_\theta}{\dd \theta}& = \frac{\partial \Gamma_\theta}{\partial\theta} + \frac{\partial \Gamma_\theta}{\partial x_T} \frac{\dd x_T}{\dd \theta} - \lambda_T^\top\frac{\dd x_T}{\dd \theta} + \lambda^\top_0\cancel{\frac{\dd x_0}{\dd\theta}}\\
            +& \int_\bT \lb\lc\frac{\partial\gamma_\theta}{\partial\theta} + \frac{\partial\gamma_\theta}{\partial x_t}\frac{\dd x_t}{\dd\theta}\rc\dd t + \dd\lambda_t^\top\frac{\dd x_t}{\dd\theta}\rb\\
            +& \int_\bT  \lambda_t^\top\lb\frac{\partial f_\theta}{\partial\theta} +\frac{\partial f_\theta}{\partial x_t}\frac{\dd x_t}{\dd\theta}\rb\dd t\\
            +& \sum_{i=1}^{\nN}\int_\bT\lambda_t^\top\lb \frac{\partial g^{(i)}_\theta}{\partial\theta} +\frac{\partial g^{(i)}_\theta}{\partial x_t}\frac{\dd x_t}{\dd\theta}\rb\circ\dd B_t^{(i)}
        \end{aligned}
    \end{equation}
    which, by reorganizing the terms, leads to 
    \begin{equation*}
        \begin{aligned}
            \frac{\dd J_\theta}{\dd \theta}& = \frac{\partial \Gamma_\theta}{\partial\theta} + \int_\bT \lb\frac{\partial \gamma_\theta}{\partial\theta}  + \lambda_t^\top\frac{\partial f_\theta}{\partial\theta}\rb\dd t\\
            & + \sum_{i=1}^\nN\int_\bT \lambda_t^\top\frac{\partial g^{(i)}_\theta}{\partial\theta} \circ\dd B_t^{(i)}            %
            +\lb\frac{\partial \Gamma_\theta}{\partial x_T}  - \lambda_T^\top\rb\frac{\dd x_T}{\dd \theta} \\
            + \int_\bT&  \lb \dd\lambda_t^\top + \left(\lambda_t^\top\frac{\partial f_\theta}{\partial x_t} + \frac{\partial \gamma_\theta}{\partial x_t}\right)\dd  t+\sum_{i=1}^\nN\lambda_t^\top\frac{\partial g^{i}_\theta}{\partial\theta} \circ\dd B_t^{i} \rb\frac{\dd x_t}{\dd\theta}
        \end{aligned}
    \end{equation*}
    Finally, if the process $\lambda_t$ satisfies the backward SDE $\dd\lambda_t^\top = -\left[\lambda_t^\top\frac{\partial f_\theta}{\partial x_t} + \frac{\partial \gamma_\theta}{\partial x_t}\right]\dd  t - \sum_{i=1}^\nN \lambda_t^\top\frac{\partial g^{(i)}_{\theta}}{\partial x_t}\circ\dd {\widecheck{B}_t^{(i)}},~~~\lambda_T^\top = \frac{\partial \Gamma_\theta}{\partial x_T}$.
    The criterion function gradient is then simply obtained by 
    \begin{equation*}
        \frac{\dd J_\theta}{\dd\theta} = \frac{\partial \Gamma_\theta}{\partial\theta}+\int_\bT\left[\lambda_t^\top\frac{\partial f_\theta}{\partial\theta} + \frac{\partial \gamma_\theta}{\partial \theta}\right]\dd  t +\sum_{i=1}^\nN\int_\bT \lambda_t^\top\frac{\partial g_{\theta}^{(i)}}{\partial\theta}\circ\dd B_t^{(i)}
    \end{equation*}
\endIEEEproof
{
Note that if the integral term of the objective function is defined point--wise at {a finite number of} time instants $t_k$, i.e. $\int_s^T\sum_{k=1}^K\gamma_\theta(t_k,\Phi_{s, t_k}^{\theta, b}(x_s))\delta(t-t_k)\dd t$, then the RHS of the adjoint state SDE becomes piece--wise continuous in $(t_k, t_{k+1}]$, yielding the \textit{hybrid} stochastic dynamics
\[
    \begin{aligned}
        &\dd\lambda_t = -\frac{\partial f_\theta}{\partial x_t}\lambda_t\dd  t - \sum_{i=1}^\nN \frac{\partial g^{(i)}_{\theta}}{\partial x_t}\lambda_t\circ\dd B_t^{(i)} && t\in(t_{k}, t_{k+1}]\\
        &\lambda_t^- = \lambda_t + \frac{\partial\gamma_\theta}{\partial x_t}&& t=t_k
    \end{aligned}
\]
where $\lambda^-_t$ indicates the value of $\lambda_t$ after a discrete \textit{jump}, { or, formally, $\lambda^-_t = \lim_{s\nearrow t^-}\lambda_s$}.
}
%
\section{Experimental Evaluation}
We evaluate the proposed approach on a classical problem in financial mathematics, continuous--time portfolio optimization \cite{merton1969lifetime}. We consider the challenging finite--horizon, transaction cost case  \cite{liu2002optimal}.
\vspace{-3mm}
\subsection{Optimal Portfolio Allocation}
Consider a two--asset market with proportional transaction costs, all securities are perfectly divisible. Suppose that in such a market all assets are traded continuously and that one of the them is \textit{riskless}, i.e. it evolves according to the ODE
\[
    \dd V_t = r_t V_t \dd t,~~~V_s = v \in\R
\]
where the \textit{interest rate} $r_t$ is any $\cF_t$-adapted non--negative scalar--valued process. The other asset is a \textit{stock} $S_t\in\R$ whose dynamics satisfy the It\^o SDEs
\[
    \dd S_{t} = \mu_{t} S_{t} + \sigma_{t}S_{t}\dd B_{t}, ~~~S_{s} = s\in\R
\]
with \textit{expected rate of return} $\mu:\bT\rightarrow\R$ and \textit{instantaneous volatility} $\sigma:\bT\rightarrow\R_+$ $\cF_t$-adapted processes.

Now, let us consider an agent who invests in such market and is interested in rebalancing a two--asset portfolio through buy and sell actions. We obtain
\begin{equation}
    \begin{aligned}
        \dd V_t &= r_t V_t \dd t - \dd \pi_t^{i} + (1 - \alpha) \dd \pi_t^{d}\\
        \dd S_{t} &= \mu_{t} S_{t} + \sigma_{t}S_{t}\dd B_{t} + \dd \pi_t^{i} - \dd \pi_t^{d}
    \end{aligned}
\end{equation}
where $\pi^{i}_t,~\pi^{d}_t$ are nondecreasing, $\cF_t$-adapted processes indicating the cumulative amount of sales and purchases of the risky asset, and $\alpha>0$ is the proportional transaction cost. Here we let $\dd\pi^{i}_t = u_t^i\dd t,~\dd\pi^{d}_t = u_t^d\dd t$ and $u_t = (u_t^i, u_t^d)$. An optimal strategy $u_t$ for such a portfolio requires specification of a utility function, encoding risk aversion and final objective of the investor. Explicit results for the finite horizon, transaction cost case exist only on specific classes of utilities, such as isoelastic \cite{liu2002optimal}. As a further complication, the horizon length itself is known to greatly affect the optimal strategy.
As an illustrative example, we consider here a portfolio optimality criteria where high levels of stock value are penalized, perhaps due to hedging considerations related to other already existing portfolios in possession of the hypothetical trader:
\[
    J_u = \int_0^T (r_t V_t + \mu_t S_t - \nu S_t^2 \sigma_t)\dd t + r_T V_T + \mu_T S_T
\]
to be maximized in expectation for some constant $\nu > 0$. The same methodology can be directly extended to cover generic objectives, including e.g different risk measures, complex stock and interest rate dynamics, and more elaborate transaction cost accounting rules. We then obtain a parameterization $u_{\theta, t}^i,~u_{\theta, t}^d$ of the policies $u^i_t, u^d_t$ as follows. In particular we define feed--forward neural networks $N_\theta:\R^2\rightarrow\R^2;~(S_t, V_t)\mapsto(u_{\theta, t}^i,u_{\theta, t}^d)$ taking as input the instantaneous values of the assets, $u_{\theta, t} = N_\theta(V_t, S_t) =\ell_L \circ \varphi \circ \ell_{L-1}\circ\cdots\circ\varphi \circ \ell_0 (V_t, S_t)$, where $\ell_k$ are linear affine finite operators, $\ell_k x = A_k x + b_k$, $A_k\in\R^{n_{k+1}\times n_k},~b_k\in\R^{n_{k +1}}$ and $\varphi:\R\rightarrow\R$ is any nonlinear \textit{activation} function thought to be acting element--wise. The vector of parameters $\theta$ is thus the flattened collection of all \textit{weights} $A_k$ and \textit{biases} $b_k$. %

\vspace{-5mm}
\subsection{Numerical Results}
In particular, we choose three--layer feed--forward neural networks with thirty--two neurons per layer, capped with a ${\tt softplus}$ activation $\varphi(x)=\log(1 + \exp(x))$ to ensure $\pi_\theta^i, \pi_\theta^d$ be nondecreasing processes.

\begin{figure}
    \centering
    \vspace{2mm}
    \includegraphics[width=\linewidth]{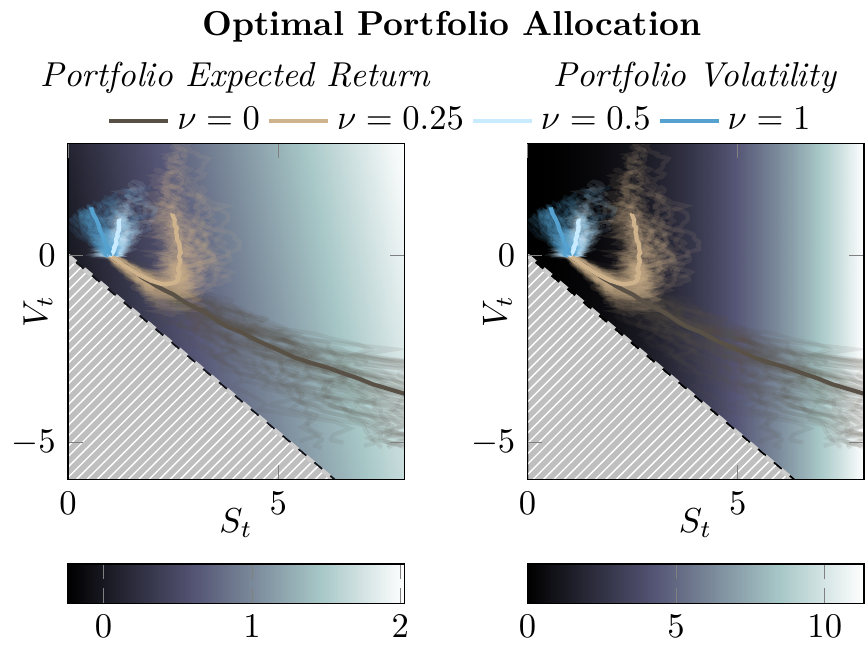}
    \vspace*{-7mm}
    \caption{Realization of $(S_t, V_t)$ trajectories under learned strategies for four investors with different degrees of risk aversion, encoded in $\nu$, with a portfolio initialized at $(1, 0)$ at time $0$. Different path realizations generated by the strategies are shown, along with their mean. [Left] Portfolio evolution against $r_t V_t + \mu_t S_t$ [Right] Portfolio evolution against $S_t^2\sigma$. Each strategy suits the corresponding risk aversion level.}
    \label{realizations}
    \vspace*{-4mm}
\end{figure}

\begin{figure*}[t]
    \centering
    \vspace{2mm}
    \includegraphics[scale=.95]{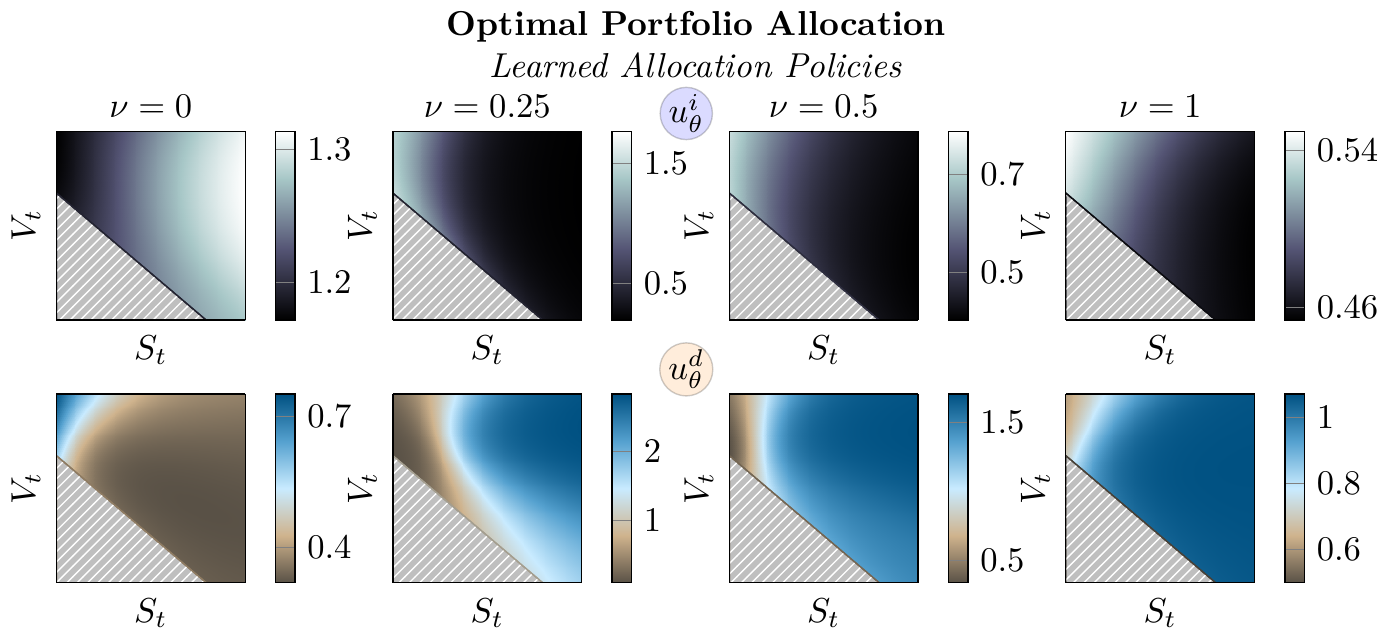}
    \vspace*{-5mm}
    \caption{Learned buy--sell strategies $u_\theta^i, u_\theta^d$ over the state--space $(S_t, V_t)$. The strategies are highly non--linear, and empirically agree with investment strategies at different risk aversion degrees. In example, the risk--seeking investor $\nu=0$ balances $u_\theta^i,u_\theta^d$ to purchase the maximum amount of stock possible while remaining above the solvency line $V_t + (1 - \alpha)S_t \geq 0$.}
    \label{strategies}
    \vspace*{-3mm}
\end{figure*}

The optimization procedure for strategy parameters $\theta$ has been carried out through $100$ iterations of the ${\tt Adam}$ \cite{kingma2014adam} gradient--based optimizer with step size $0.03$. The asset dynamics and cost parameters are set as $\alpha = 0.05$, $r = 0.04$, $\mu=0.23$, $\sigma = 0.18$. We produce $50$ path realizations during each training iteration and average the gradients. The asset dynamics are interpreted in the It\^o sense and have been solved forward with the It\^o--Milstein method \cite{mil1975approximate}, which has been converted to Stratonovich--Milstein for the backward adjoint SDE. Figure \ref{realizations} shows different realization of portfolio trajectories in the $(S_t, V_t)$ state--space for four different values of $\nu$, $[0, 0.25, 0.5, 1]$. The trajectories agree with plausible strategies for each investment style; risk--seeking investors ($\nu=0$) sell $V_t$ to maximize the amount of stock owned at final time $T$, and hence potential portfolio growth. On the other hand, moderately risk--averse investors ($\nu=0.25$) learn hybrid strategies where an initial phase of stock purchasing is followed by a movement towards a more conservative portfolio with a balanced amount of $S_t$ and $V_t$. The policies $u_\theta^i, u_\theta^d$ are visualized in Figure \ref{strategies}, confirming the highly non--linear nature of the learned controller. It should be noted that for a strategy to be allowed, the portfolio has to remain above the solvency line $V_t + (1 - \alpha)S_t \geq 0$  \cite{liu2002optimal}, indicated in the plot. We empirically observe automatic satisfaction of this specific constraint for all but the investors with lowest risk aversion. For the case $\nu=0$, we introduce a simple logarithmic barrier function as a soft constraint \cite{nocedal2006numerical}.
%
\section{Related Work}
Computing path--wise sensitivities through stochastic differential equations has been extensively explored in the literature. In particular, \textit{forward} sensitivities with respect to initial conditions of solutions analogous to Proposition \ref{thm:1} were first introduced in \cite{kunita1997stochastic} and extended for It\^o case to parameters sensitivities in \cite{gobet2005sensitivity}. These approaches rely on integrating forward matrix--valued SDEs whose drift and diffusion require full Jacobians of $f_\theta$ and $g_\theta$ at each integration step. Thus, such methods poorly scales in computational cost with large number of parameters and high--dimensional--state regimes. This issue is averted by using \textit{backward} adjoint sensitivity \cite{giles2005smoking} where a vector--valued SDE is integrated backward and only requires vector--Jacobian products to be evaluated. In this direction, \cite{li2020scalable} proposed to solve the system's backward SDE alongside the adjoint SDE to recover the value of the state $x_t$ and thus improve the algorithm memory footprint. Further, the approach is derived as a special case of \cite[Theorem 2.4.1]{kunita2019stochastic} and only considers criteria $J_\theta$ depending on the final state $x_T$ of the system.%

Our stochastic adjoint process extends the results of \cite{li2020scalable} to \textit{integral} criteria potentially exhibiting explicit parameter dependence. Further, we proposed a novel proving strategy based on classic variational analysis.%





%
\bibliographystyle{IEEETran}
\bibliography{main}
\end{document}